\documentclass{article}
\pdfoutput=1
\usepackage{iclr2016_conference}
\usepackage[utf8]{inputenc}
\usepackage{times}
\usepackage{graphicx}
\usepackage{subcaption}

\usepackage{natbib}
\usepackage{algorithm}
\usepackage{algorithmic}
\usepackage{amsmath}
\usepackage{amsfonts} 
\usepackage{amssymb}
\usepackage{bm}
\allowdisplaybreaks[0]
\usepackage{enumitem}
\usepackage{microtype}
\usepackage{xargs}
\usepackage{booktabs}

\definecolor{OliveGreen}{rgb}{0.33, 0.42, 0.18}
\definecolor{Plum}{rgb}{0.56, 0.27, 0.52}

\usepackage[colorinlistoftodos,prependcaption,textsize=tiny]{todonotes}  
\newcommandx{\unsure}[2][1=]{\todo[linecolor=red,backgroundcolor=red!25,bordercolor=red,#1]{#2}}
\newcommandx{\change}[2][1=]{\todo[linecolor=blue,backgroundcolor=blue!25,bordercolor=blue,#1]{#2}}
\newcommandx{\info}[2][1=]{\todo[linecolor=OliveGreen,backgroundcolor=OliveGreen!25,bordercolor=OliveGreen,#1]{#2}}
\newcommandx{\improvement}[2][1=]{\todo[linecolor=Plum,backgroundcolor=Plum!25,bordercolor=Plum,#1]{#2}}

\PassOptionsToPackage{hyphens}{url}
\usepackage[colorlinks]{hyperref}
\definecolor{mydarkblue}{rgb}{0,0.08,0.45}
\hypersetup{ %
    pdftitle={},
    pdfauthor={},
    pdfsubject={},
    pdfkeywords={},
    pdfborder=0 0 0,
    pdfpagemode=UseNone,
    colorlinks=true,
    linkcolor=mydarkblue,
    citecolor=mydarkblue,
    filecolor=mydarkblue,
    urlcolor=mydarkblue,
    pdfview=FitH}

\usepackage[english]{babel}
\addto\extrasenglish{
}

\newcommand\ddfrac[2]{\frac{\displaystyle #1}{\displaystyle #2}}

\DeclareMathOperator*{\argmax}{\arg\!\max}

\title{Binding via Reconstruction Clustering}

\author{Klaus Greff, Rupesh Kumar Srivastava \& Jürgen Schmidhuber\\
The Swiss AI lab IDSIA, USI-SUPSI\\
Lugano, Switzerland\\
\texttt{\{klaus,rupesh,juergen\}@idsia.ch}
}

\begin{document}

\maketitle


\begin{abstract}
Disentangled distributed representations of data are desirable for machine learning, since they are more expressive and can generalize from fewer examples.
However, for complex data, the distributed representations of multiple objects present in the same input can interfere and lead to ambiguities, which is commonly referred to as the \emph{binding problem}. 
We argue for the importance of the binding problem to the field of representation learning, and develop a probabilistic framework that explicitly models inputs as a composition of multiple objects.
We propose an algorithm that uses a denoising autoencoder to dynamically bind features together in multi-object inputs through an Expectation-Maximization-like clustering process. 
The effectiveness of this method is demonstrated on artificially generated datasets of binary images, showing that it can even generalize to bind together new objects never seen by the autoencoder during training.
\end{abstract}


\section{The Binding Problem}
Two important properties of good representations are that they are \emph{distributed} and \emph{disentangled}.
Distributed representations~\citep{Hinton1984} are far more expressive than local ones, requiring exponentially fewer features to capture the same space.
Complementary to that, disentangling~\citep{Barlow1989, Schmidhuber1992, Bengio2007} requires the factors of variation in the data to be separated into different independent features.
This concept is closely related to invariance and eases further processing because many properties, that we might be interested in, are invariant under a wide variety of transformations~\citep{Bengio2013a}.
Unfortunately distributed representations can interfere and lead to ambiguities when multiple objects are to be represented at the same time.

\begin{figure}[t]
    \centering
    \begin{subfigure}[b]{0.3\textwidth}
        \centering
        \includegraphics[height=0.15\textheight]{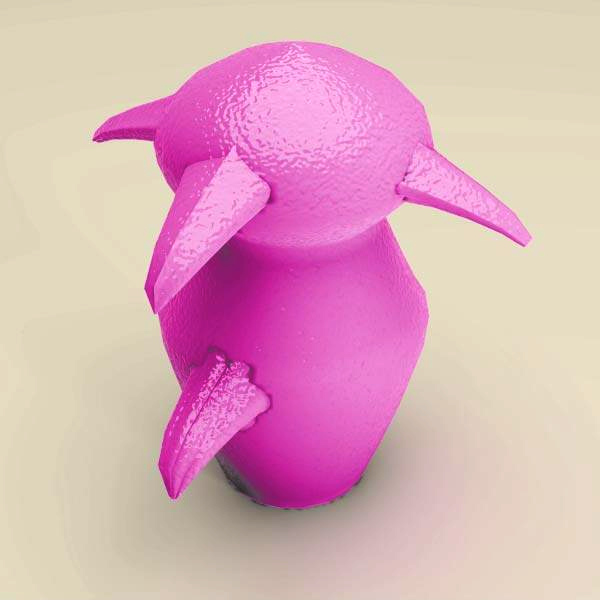}
        \caption{}
        \label{fig:greeble}
    \end{subfigure}
    \hfill
    \begin{subfigure}[b]{0.3\textwidth}
        \centering
        \includegraphics[height=0.15\textheight]{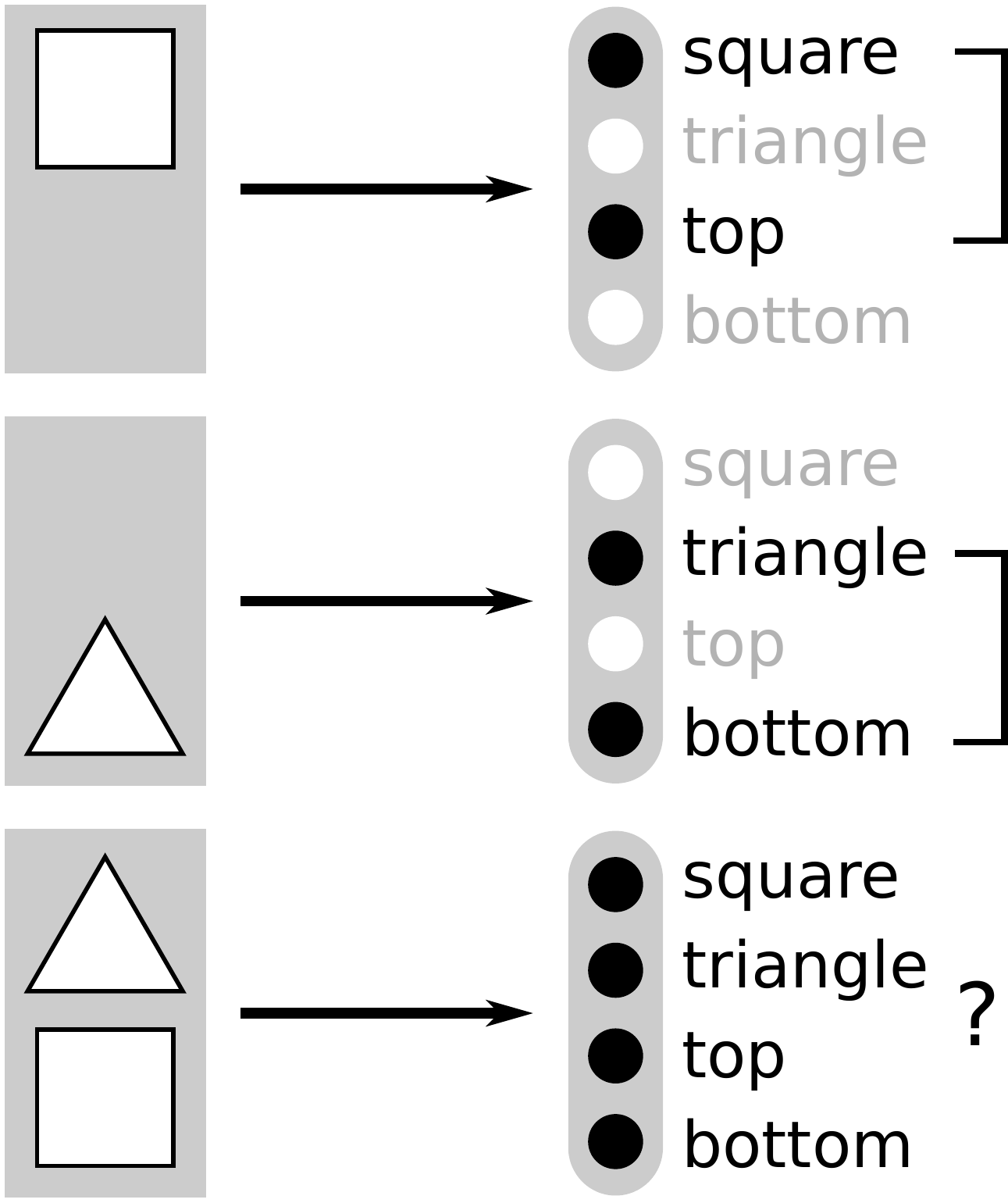}
        \caption{}
        \label{fig:interference}
    \end{subfigure}
    \hfill
    \begin{subfigure}[b]{0.3\textwidth}
        \centering
        \includegraphics[height=0.15\textheight]{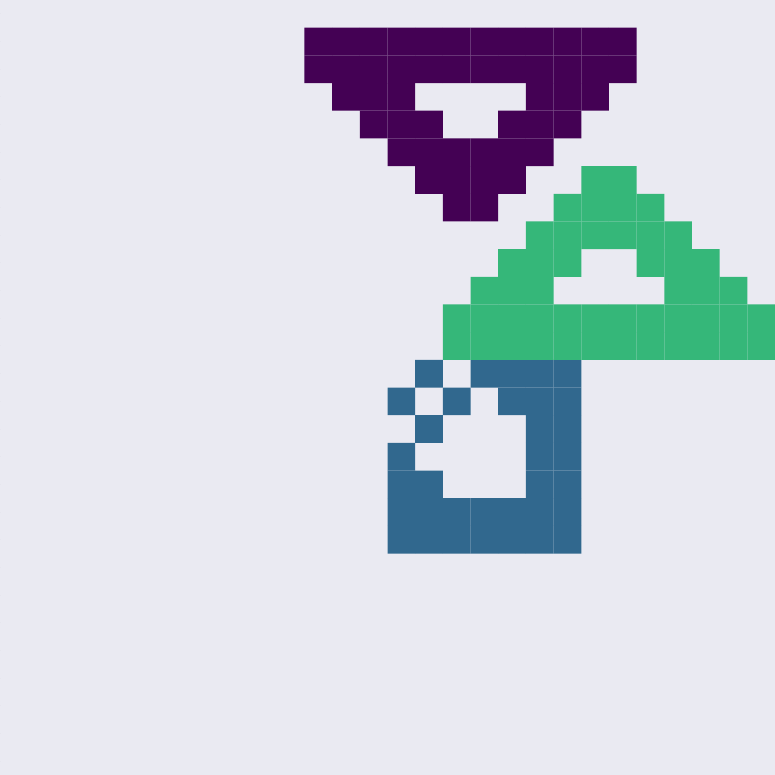}
        \caption{}
        \label{fig:obj_pred}
    \end{subfigure}
    \caption{(a) A Greeble. (b) Example demonstrating the binding problem (c) An illustration of intra-object predictability. The missing pixels from the square can be predicted using other pixels constituting the box, but not from pixels constituting other objects.}
\end{figure}

The \emph{binding problem} refers to these ambiguities that can arise from the superposition of multiple distributed representations.
This problem has been debated quite extensively in the neuroscience and psychology communities perhaps starting with \citet{milner1974} and \citet{vonderMalsburg1981}, but its existence can be traced back at least to a description by \citet{rosenblatt1961}. 
It is classically demonstrated with a system required to identify an input as either square($\square$) or triangle($\triangle$) and to decide whether it is at the top($\uparrow$) or at the bottom($\downarrow$). 
It represents every object as a distributed representation with two active disentangled features (see \autoref{fig:interference}). 
The binding problem arises when the system is presented with two objects at the same time: 
In this scenario, all four features become active and from the representation alone it cannot be determined whether the input contains a square on top and a triangle at the bottom or vice versa.

One way the system can circumvent this problem is through the use of a \emph{local} representation, with one feature for each combination of shape and position: $\triangle_\uparrow, \triangle_\downarrow, \square_\uparrow, \square_\downarrow$.
Sadly the size of such a purely local-representation scales exponentially with the number of factors to represent.
In contrast, \emph{distributed representations} \citep{Hinton1984} are much more expressive and can generalize better through the reuse of features. 
The former system could, for example, correctly represent the position for a new object such as a circle by the already available position features ($\uparrow\downarrow$).

Generalization of internal representations is a crucial capability of any intelligent system, and one that still sets humans apart from current machine learning systems. 
Consider \autoref{fig:greeble}, an example from studies in psychology:
chances are this is the first time you see a \emph{Greeble}~\citep{gauthier1997}.
Nevertheless, you are capable of describing its shape, texture, and color. 
Moreover, you can easily segment it and tell it apart from the background, without having seen any other Greeble before.
It has long been argued that such generalization capabilities are a result of the use of distributed representations in the human brain.

Despite its importance to the neuroscience community, binding has received relatively little attention in representation learning. 
Two important reasons for this are:

Firstly, most pattern recognition tasks and benchmarks are set up to avoid the binding problem. 
Many popular visual pattern recognition datasets consist of images that contain only one object at a time.
Similarly, speech recognition mostly considers recordings of just one speaker talking and little background noise.
In these settings, a machine learning algorithm can assume that there is only a single prominent object of interest, reducing the binding problem to the problem of ignoring irrelevant details. 
When tackling more challenging problems such as image caption generation, scene parsing segmentation, or the cocktail party problem, the deficiencies of popular methods become more apparent and restrictive.

Secondly, the recent increase in processing power due to the use of Graphics Processing Units made it feasible to mitigate the binding problem using localized binding in the form of convolutions~\citep{riesenhuber1999a}.
Convolutional Networks~\citep{Fukushima1979, lecun1990} use feature detectors with limited receptive fields (filters) replicated over the whole input to represent its inputs. 
Therefore the resulting features of spatially separated objects do not interact: they are invariant to changes outside their field of view.
On the other hand, they do not disentangle the location from the detected pattern, which comes at the cost of having to compute the same feature replicated many times over the image.
While this is reasonable for low-level features like edges, it seems wasteful to replicate specialized high-level features such as dog-faces. 

In this paper, we develop an unsupervised method that \emph{dynamically} binds features of different objects together.
This is in contrast to local representations which by nature \emph{statically} bind several input features together (a feature for $\square_\uparrow$ permanently binds the concepts $\square$ and $\uparrow$ together).
It explicitly models inputs as a composition of multiple entities and recovers these ``objects" using the notion of mutual predictability.
This is achieved through a clustering process which utilizes a denoising autoencoder (DAE; \citealp{behnke2001, vincent2008}) to iteratively reconstruct an input.
In the future, such a mechanism could help to effectively use distributed representations in multi-object settings without being impaired by the ambiguities due to superposition.
Alternative approaches to the binding problem proposed in the literature are discussed in \autoref{sec:related}.

\section{Reconstruction Clustering}
This section describes \emph{Reconstruction Clustering} (RC), a formal framework for tackling the binding problem as a clustering problem.
For ease of explanation we will refer to inputs as images and the individual dimensions of an input as pixels, though the framework is not restricted to visual inputs.
It is based upon two insights: Firstly, if the image was segmented into its constituent objects, there would be no binding problem. Secondly, the intuitive notion of an object can be formalized as a group of mutually predictive pixels.
The proposed method therefore iteratively clusters pixels based on how well they predict each other.

\subsection{Images as Compositions}
The first central idea behind RC is to model images as being composed of several independent objects with each pixel belonging to one of them.
Unlike in classic segmentation where each pixel is assigned to a predefined class, the goal here is to simply segregate different objects. 
In doing so we avoid all ambiguities that might arise from a superposition of their representations.
Of course, the information about which objects are present and which pixels they consist of is unknown in practical applications.  
So for each image, the aim is to infer both the object representations and the corresponding pixel assignments.

\begin{figure}
    \centering
    \begin{subfigure}[c]{0.3\textwidth}
        \centering
        \includegraphics[height=0.2\textheight]{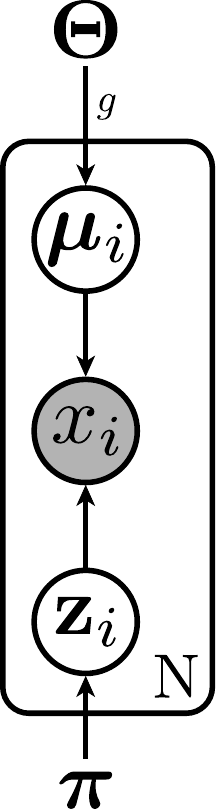}
        \caption{}
        \label{fig:prob_struct}
    \end{subfigure}
    \hfill
    \begin{subfigure}[c]{0.69\textwidth}
        \centering
        \includegraphics[height=0.3\textheight]{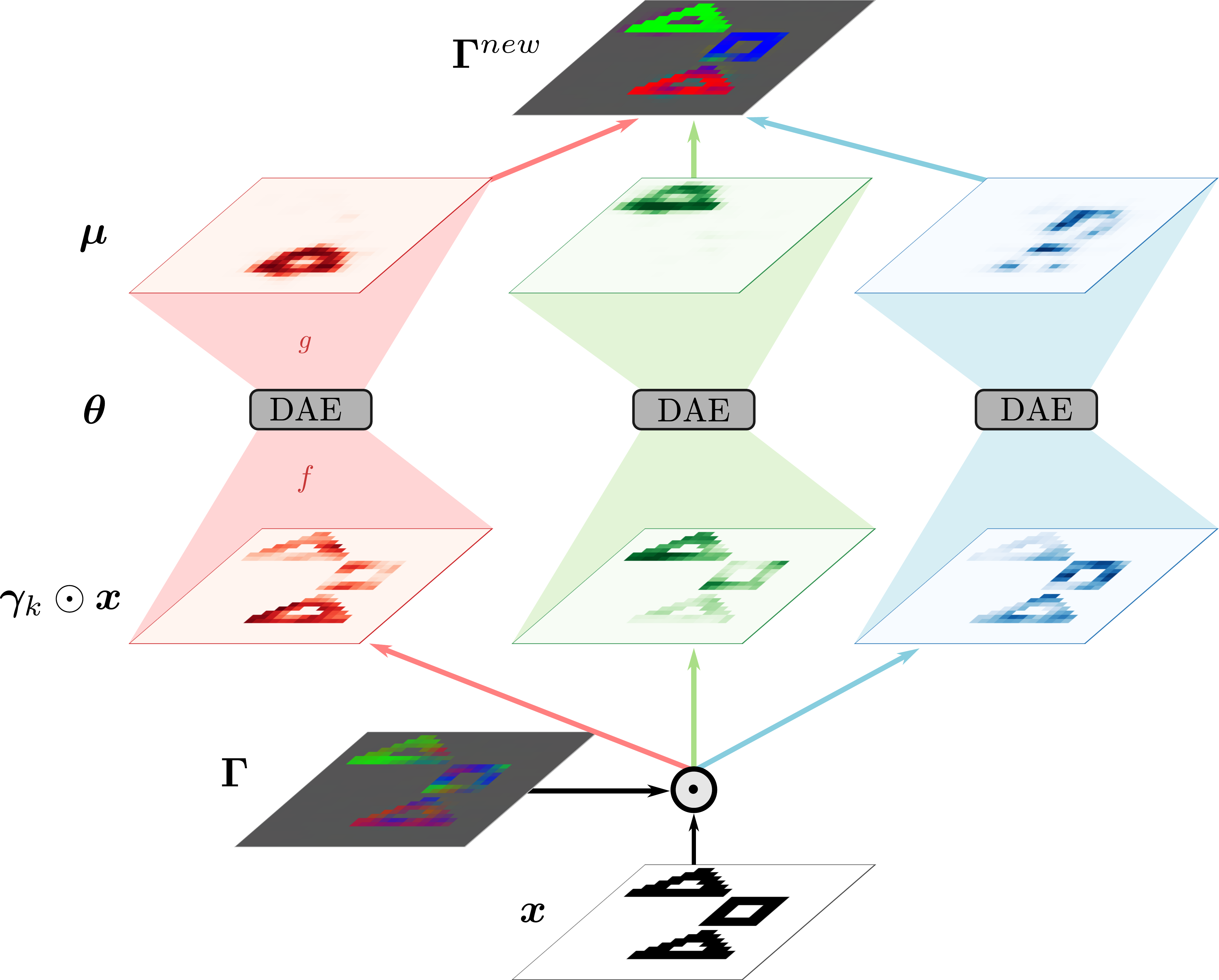}
        \caption{}
    \label{fig:alg_illustr}
    \end{subfigure}
    \caption{(a) The assumed probabilistic structure. (b) A schematic illustration of one iteration of the RC algorithm.}
\end{figure}

Formally, we introduce a binary latent vector $\mathbf{z_i}$ for each pixel $x_i$ that specifies which of the $K$ objects it belongs to. 
Therefore, $\mathbf{z_i} = (z_{i1}, z_{i2}, \dots, z_{iK}) \in \{0, 1\}^K$ with the constraint that $\sum_{k=1}^K z_{ik} = 1$. 
Let $N\in\mathbb{N}$ denote the number of pixels in the image $\mathbf{x} = \{x_1, \dots, x_N \}$.
Then we define the prior over $\mathbf{Z}=\{\mathbf{z}_1, \dots, \mathbf{z}_N\}$ as:

\begin{equation}
    P(\mathbf{Z}|\bm{\pi}) = \prod_{i=1}^N P(\mathbf{z}_i|\bm{\pi}) = \prod_{i=1}^N \prod_{k=1}^K \pi_k^{z_{ik}},
\label{eq:z_prior}
\end{equation}

where the $\mathbf{z}_i$'s are assumed to be independent given $\bm{\pi} = \{\pi_1, \dots, \pi_K\}$.
The assumed probabilistic structure is shown in \autoref{fig:prob_struct}.
We assume $\bm{\pi}$ to be uniformly distributed for simplicity, but its estimation can be incorporated into the algorithm if required (see Appendix). 

\subsection{Objects}
So far, we have used the word \emph{object} to describe a group of pixels that somehow ``belong together".
The second central idea of RC is to concretize that notion using mutual predictability of the pixels.
Intuitively, knowing about some pixel values that belong to an object helps in predicting the others. 
An example can be seen in \autoref{fig:obj_pred} where the corrupted pixels in the bottom left corner of the square could be reconstructed from knowledge about the rest of the square, but not from any of the triangles.
So we define an object as a group of pixels that help in predicting each other, but do not carry information about pixels outside of that group.

Predictability, as we use it here, is derived from the structure of the underlying data-distribution.
Knowledge about this structure is also precisely what is needed in order to remove corruption from an image.
Based on this insight, we propose to use a denoising autoencoder (DAE) to measure predictability.

\subsection{Denoising Autoencoder}
Let $f$ be the encoder and $g$ be the decoder of a DAE, such that $\bm{\theta} = f(\mathbf{x})$ is the encoded representation of input $\mathbf{x}$ and $\bm{\mu} = g(\bm{\theta})$ is the decoded output. 
The DAE is trained to remove corruption from images of single objects and thus learns a local model of the data generating distribution~\citep{vincent2008,bengio2013}.
After training the same DAE is used for each of the clusters to get predictions $\mu_{ik}$ from cluster $k$ for pixel $i$, where the object in cluster $k$ is represented by $\bm{\theta}_k$:
\begin{equation}
    P(\mathbf{x}|\bm{\theta}_k) = \prod_{i=1}^N P(x_i|\bm{\theta}_k) = \prod_{i=1}^N P(x_i|\mu_{ik})
\label{eq:x_given_theta}
\end{equation}

Here $x_i$'s are assumed to be independent given $\bm{\mu}$. Combining this with the latent variables we get:
\begin{equation}
    P(\mathbf{x}|\mathbf{Z}, \theta) = \prod_{i=1}^N \prod_{k=1}^K P(x_i|\mu_{ik})^{z_{ik}}
\label{eq:x_given_z_theta}
\end{equation}

\subsection{Clustering}

We can now outline a clustering algorithm that estimates the object identities and the corresponding pixel assignments.
Formally, we seek to maximize the complete data log-likelihood:

\begin{equation}
    \log P(\mathbf{x}, \mathbf{Z}| \bm{\mu}, \bm{\pi}) = 
        \sum_{i=1}^N \sum_{k=1}^K z_{ik} (\log P(x_i|\mu_i) + \log \pi_k)
\label{eq:log_l}
\end{equation}

This can be done in an iterative procedure where we start by randomly initializing the latent cluster assignments $\mathbf{Z}$ and then alternating between the following two steps:

\begin{enumerate}
    \item Apply the autoencoder to the each of the $K$ images that are assigned to the clusters to get a new estimate of the $K$ object representations. (R-step)
    \item Re-assign the pixels to the clusters according to their reconstruction accuracy. (E-step)
\end{enumerate}

\subsubsection{Reconstruction Step}
The R-step applies the encoder to generate a new object representation from each of the $K$ partial images that are assigned to each cluster.
We call this representation of a partial image since the encoder only gets to see as much of each pixel of the original image as has been soft-assigned to the current cluster.
The DAE then denoises the ``corruption'' caused by the cluster assignments.
The R-Step is thus given by the following formula, where $\odot$ denotes point-wise multiplication:
\begin{equation}
    \bm{\theta}_k = f(\bm{\gamma}_k \odot \mathbf{x}),
\label{eq:m_step}
\end{equation}

Unfortunately this step can not be guaranteed to increase the expected log-likelihood, because only in expectation does the DAE map from regions of low likelihood to regions of higher likelihood. 
Moreover, this property only holds for the whole image and not for all subsets of pixels.
Thus, convergence can't be proven and RC is not an Expectation Maximization algorithm \citep{Dempster1977}.
Nevertheless, empirical results show that convergence does occur reliably (\autoref{sec:convergence}). 

\subsubsection{Estimation Step}
In the E-step, for each pixel $x_i$ the posterior $\gamma_{ik}$ of $\mathbf{Z}$, given the data and the predictions $\bm{\mu}_i = \{g(\bm{\theta}_1)_i, \dots, g(\bm{\theta}_K)_i\}$ of the autoencoders based on the object representations, is
\begin{equation}
    \gamma_{ik} = P(z_{ik}=1|x_i, \bm{\mu_i}, \bm{\pi}) = \ddfrac
         {P(x_i|z_{ik}=1, \bm{\mu_i}) P(z_{ik}=1|\bm{\pi})}
         {P(x_i|\bm{\mu_i}, \bm{\pi})}.
\label{eq:gamma1}
\end{equation}

In this paper, we assume the pixels $\mathbf{x}$ to be binary and the predictions of the network $\mu$ to correspond to the mean of a binomial distribution.
Then the following performs a soft-assignment of the pixels to the $K$ different clusters:
\begin{equation}
    \gamma_{ik} = \ddfrac
        {\mu_{ik}^{x_i} (1-\mu_{ik})^{1-x_i} \pi_k }
        {\sum_{j=1}^K \mu_{ij}^{x_i} (1-\mu_{ij})^{1-x_i} \pi_j }.
\label{eq:gamma2}
\end{equation}


\section{Experiments}
\begin{figure}[t]
\centering
    \includegraphics[width=\textwidth]{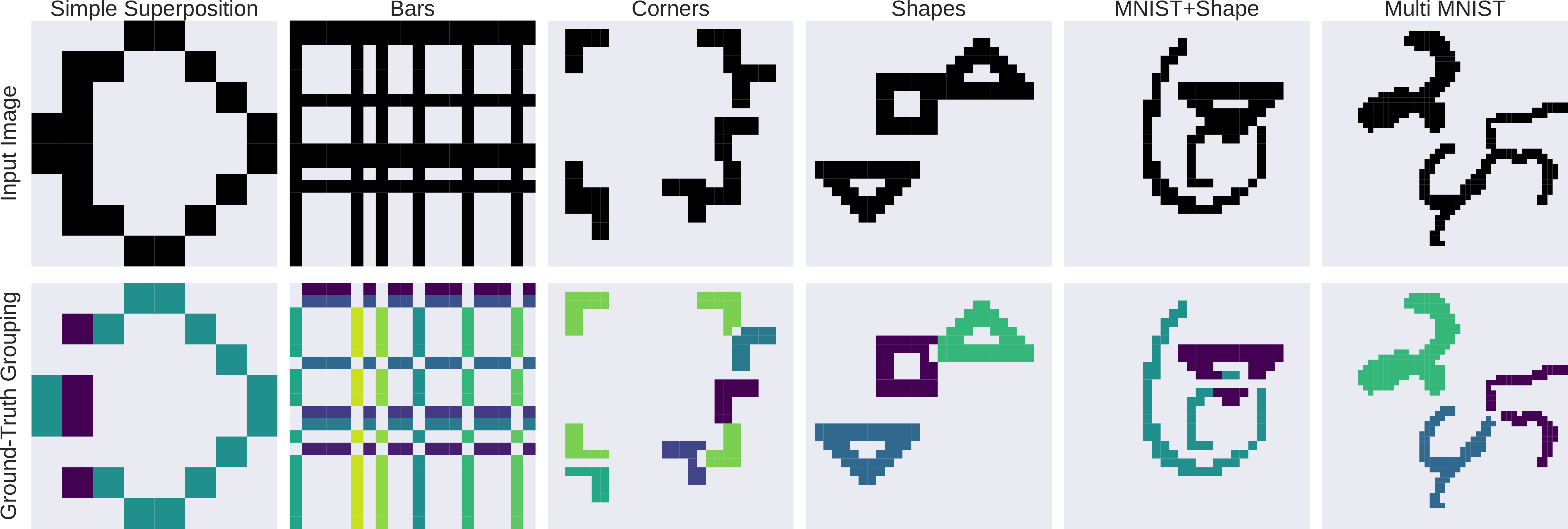}
    \caption{One example from each of the six datasets. The input images are shown on the top row with the corresponding ground-truth grouping below.}
\label{fig:data_preview}
\end{figure}

We evaluated RC on a series of artificially generated datasets consisting of binary images of varying complexity.
For each dataset, a DAE was trained to remove salt\&pepper noise on images with single objects.
The autoencoders used were fully-connected feed-forward neural networks with a single hidden layer and sigmoid output units.
A random search was used to select appropriate hyperparameters (see Appendix for details).
The best DAE obtained for each dataset was used for reconstruction clustering on 1000 test images containing multiple objects, and the binding performance was evaluated based on groud-truth object identities. All the code for this paper (including the creation of the datasets and figures) is available online on \href{https://github.com/Qwlouse/Binding}{GitHub.com/Qwlouse/Binding}.

\subsection{Datasets}
Representative examples from the datasets are shown in \autoref{fig:data_preview}.
\begin{description}
\item[Simple Superposition] A collection of simple pixel patterns two of which are superimposed. Taken from \cite{Rao2008}. This is a simple dataset with no translations, but significant overlap between patterns. 
\item[Shapes] Taken from \cite{Reichert2013}. Three shapes $(\square,\triangle,\bigtriangledown)$ are randomly placed in an image (possibly with overlap). This dataset tests binding of shapes under translation invariance and varying overlap.
\item[Bars] Introduced by \cite{foeldiak1990} to demonstrate unsupervised learning of independent components of an image. We use the variant from \cite{Reichert2013} which employs 6 horizontal, and 6 vertical lines placed in random positions in the image. 
\item[Corners] This dataset consists of 8 corner shapes placed in random orientations and positions, such that 4 of them align to form a square. It was introduced by \cite{Reichert2013} to demonstrate that spatial connected-ness is not a requirement for binding.
\item[MNIST+Shape] Another dataset from \cite{Reichert2013}, which combines a random shape from the shapes dataset with a single MNIST digit. This dataset is useful to investigate binding multiple types of objects.
\item[Multi-MNIST] Three random MNIST digits are randomly placed in a 48$\times$48 image. It provides a more challenging setup with multiple complex objects.
\end{description}

\subsection{Evaluation}
Since the data is generated, a ground-truth segmentation for each image is available. 
For the binding task, all pixels corresponding to the same object should be clustered together.
We evaluated performance by measuring the Adjusted Mutual Information (AMI; \citealp{Vinh2010}) between the true segmentation and the result of the binding, to which we refer to as the \emph{score}.
This score measures how well two cluster assignments agree and takes a value of 1 when they are equivalent, and 0 when their agreement corresponds to that expected by chance.
Only pixels that unambiguously belong to one object were counted, ignoring background pixels and regions where multiple objects overlap.

\section{Results}
\subsection{Scores}
\begin{figure}
    \centering
    \begin{subfigure}[b]{0.49\textwidth}
        \centering
        \includegraphics[height=0.19\textheight]{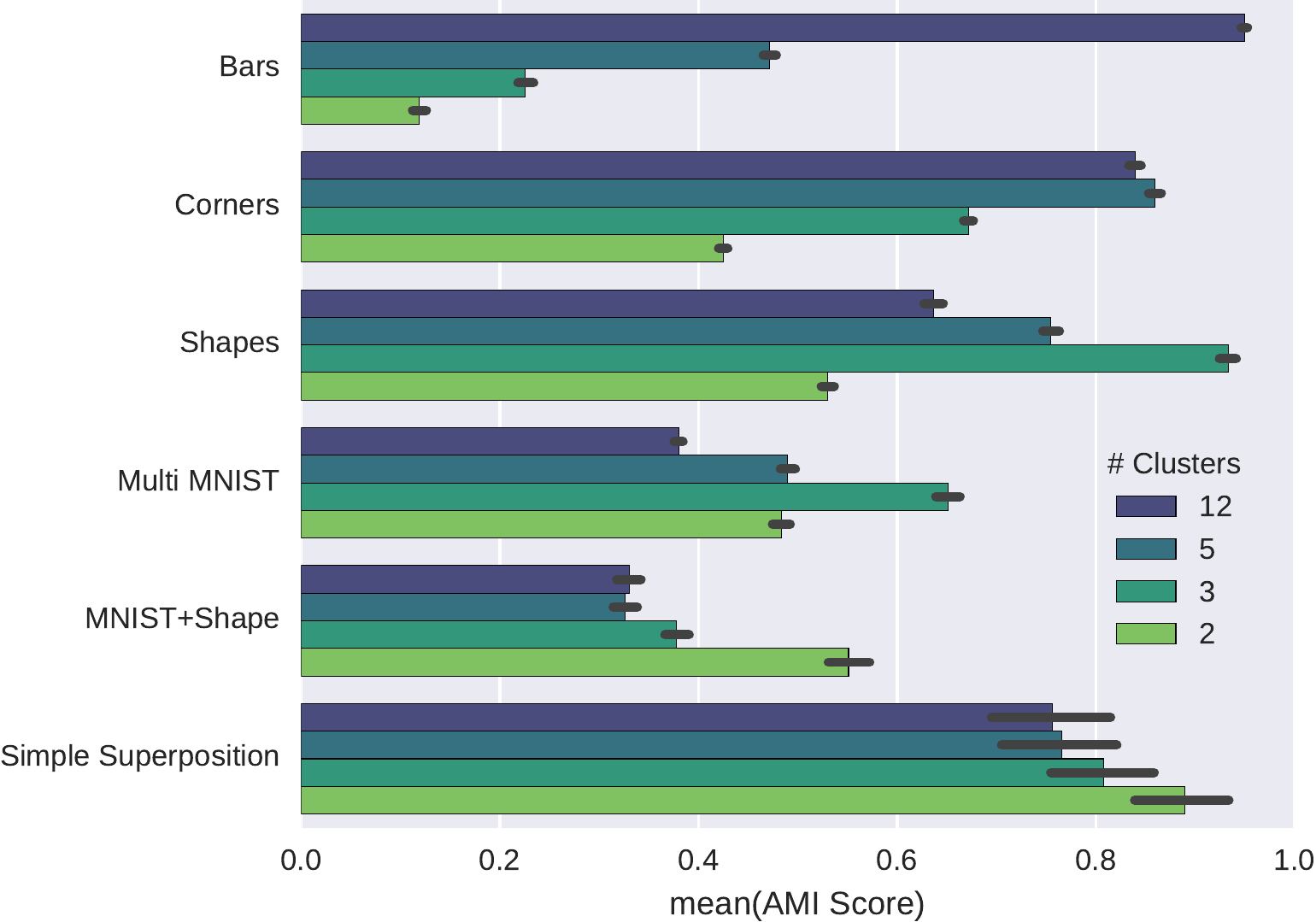}
        \caption{Overall Scores}
        \label{fig:results}
    \end{subfigure}    
    \hfill
    \begin{subfigure}[b]{0.49\textwidth}
        \centering
        \includegraphics[height=0.19\textheight]{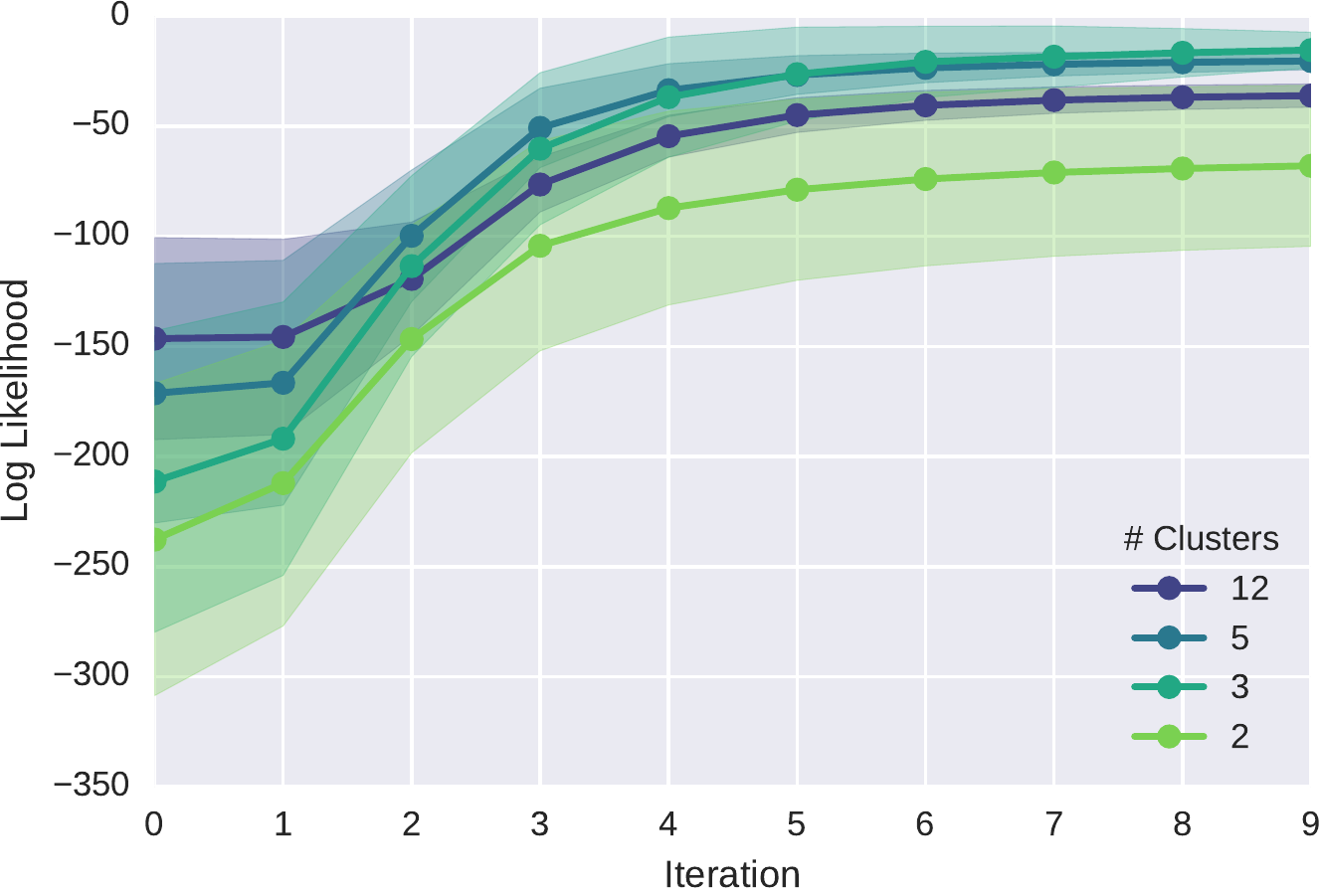}
        \caption{Convergence}
        \label{fig:convergence}
    \end{subfigure}
    \caption{
    \textbf{Left:} Mean AMI score over 1000 test samples for all datasets and various number of clusters $K$.
    \textbf{Right:} Convergence of the log-likelihood on the \emph{shapes} dataset for different numbers of clusters, showing test set mean (line) and standard deviation (shaded) over the test set.}
\end{figure}

\autoref{fig:results} shows the mean scores obtained using RC for each dataset averaged over 100 runs. Scores obtained with different choices of the number of clusters $K$.
Results are consistent across runs, hence the standard deviations are very low and barely visible.
The optimal number of clusters is two for \emph{Simple Superposition} and \emph{MNIST+Shape}, three for \emph{Multi MNIST} and \emph{Shapes}, five for \emph{Corners}, and 12 for \emph{Bars}.
Scores are higher than 0.5 for all datasets and higher than 0.8 for four out of the six datasets demonstrating the ability of RC to successfully bind objects together. 

\subsection{Convergence}\label{sec:convergence}
\autoref{fig:convergence} shows the convergence of the mean log-likelihood over RC iterations on the \emph{shapes} dataset.
Convergence is quick, typically within 5-10 iterations, depending on the chosen number of clusters $K$ and the dataset (not shown).
As expected, the final likelihood is highest when the number of clusters equals the number of objects in the shapes dataset (3), matching the results from \autoref{fig:results}. 
The likelihood is much lower for $k=2$ than for $k=3$ and drops again slightly if we choose $k=5$.
The likelihood for $k=12$ is significantly lower.
In some cases the correct choice of $k$ did not result in the highest likelihood, but in general this correspondence appeared to hold. 
If the number of objects is unknown, this trend can be used to determine the correct number of clusters.

\subsection{Qualitative Analysis}
\begin{figure}
\centering
\includegraphics[width=\textwidth]{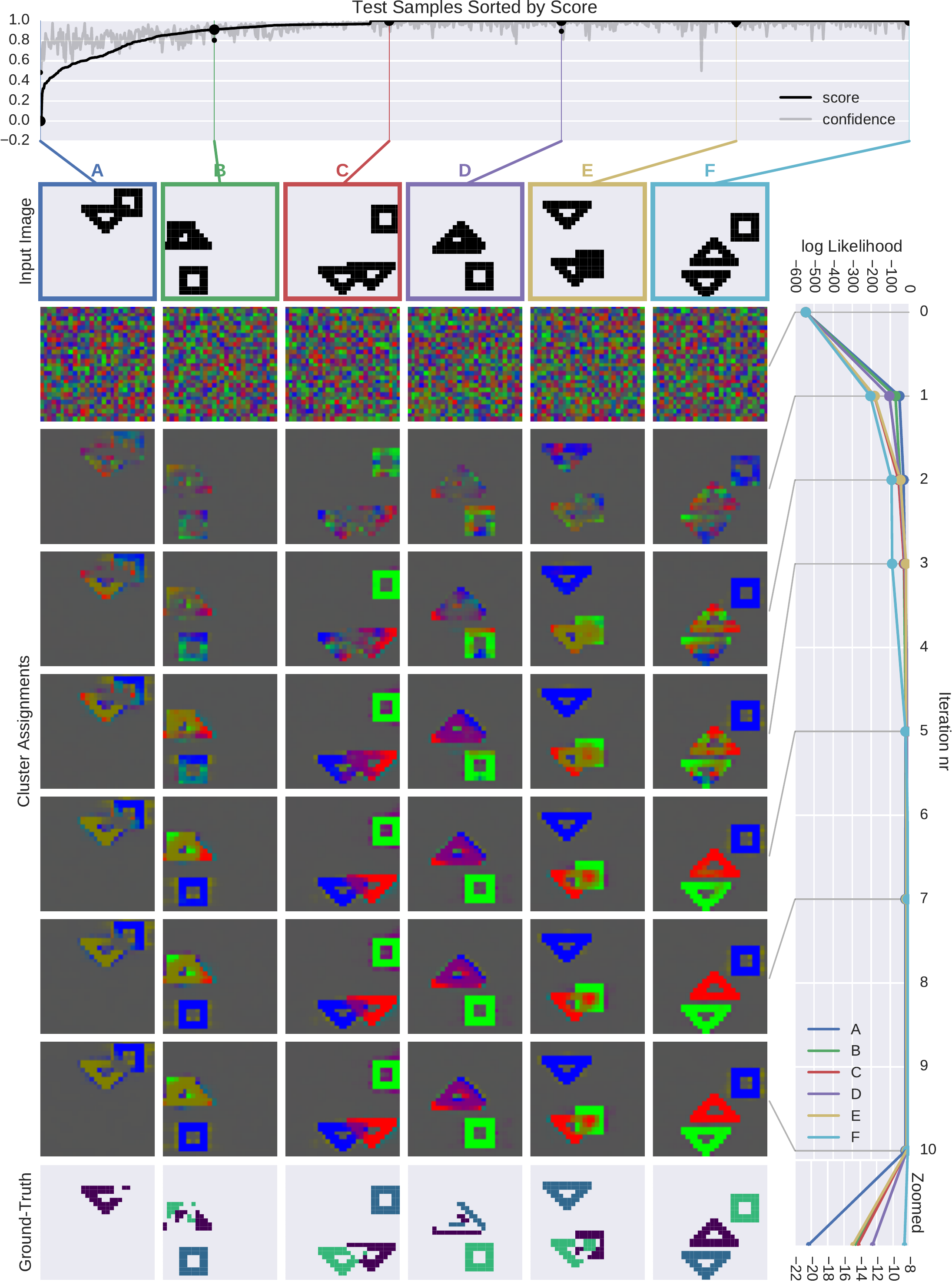}
\caption{The top plot shows the score and confidence for each of the 1000 test images from the shapes dataset, sorted by score. 
The confidence is the average value of $\max_k \gamma_{ik}$ for each evaluated pixel  (non-background, non-overlap). 
The central part of the figure shows six examples (columns) along with the cluster assignments (indicated by different colors) over RC iterations.
The corresponding ground-truth is shown at the bottom.
The right vertical plot shows the log-likelihood over the RC iterations corresponding to the displayed cluster assignments.
Similar plots for the other datasets are included in the Appendix.}
\label{fig:shapes_iter}
\end{figure}

\autoref{fig:shapes_iter} shows a few example RC runs of on the shapes dataset for qualitative evaluation.
The initial cluster assignments are random, therefore all observed structure is due to the clustering process.
The final clustering corresponds well to the ground truth even for cases with significant overlap. Again, it is notable that RC converges quickly (within 5 iterations).

\subsection{Loss vs Score}
\begin{figure}
\centering
    \begin{subfigure}[b]{0.6\textwidth}
    \centering
        \includegraphics[height=0.2\textheight]{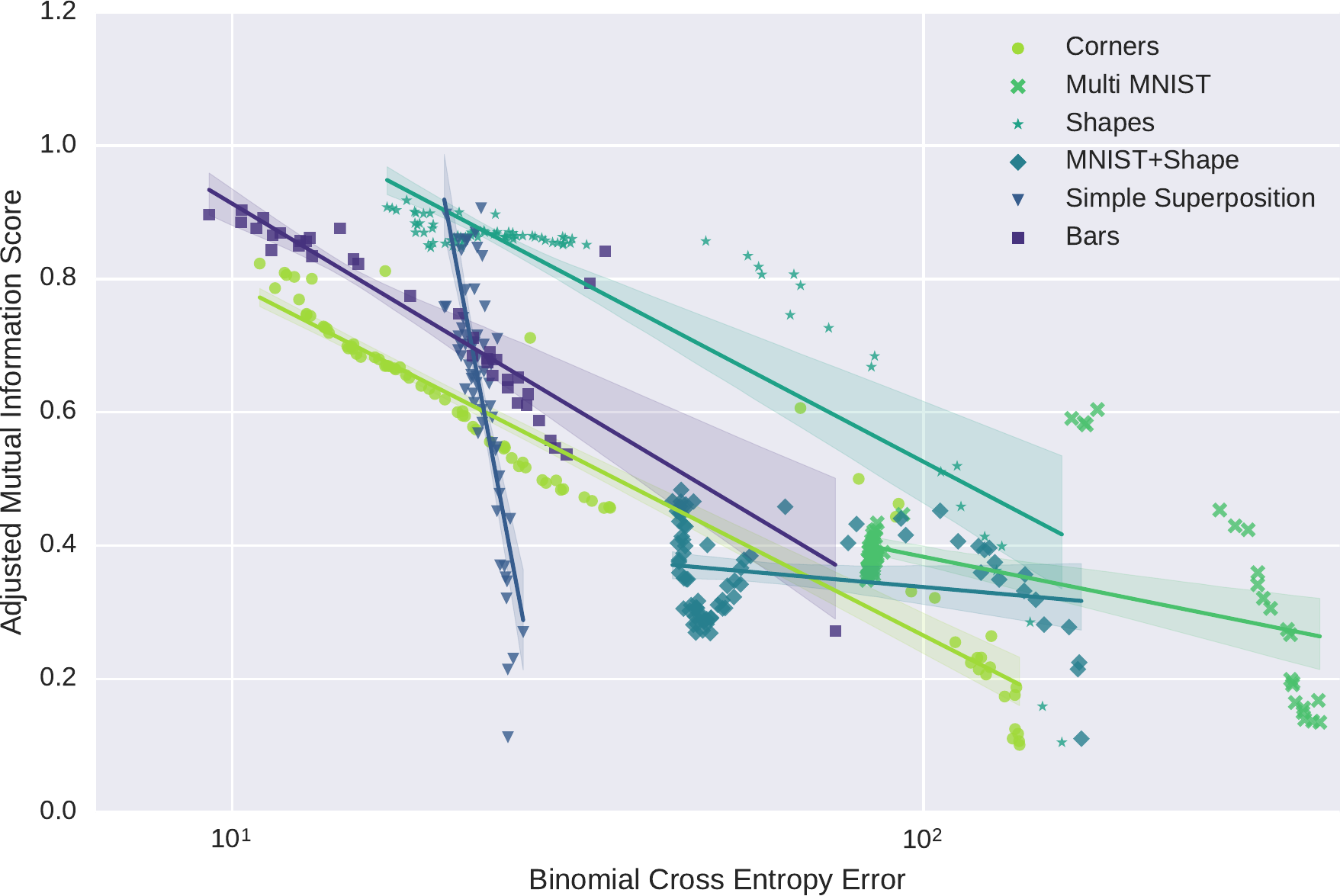}
        \caption{Loss Vs Score}
    \label{fig:score_vs_loss}
    \end{subfigure}    
    \hfill
    \begin{subfigure}[b]{0.38\textwidth}
        \centering
    \includegraphics[height=0.2\textheight]{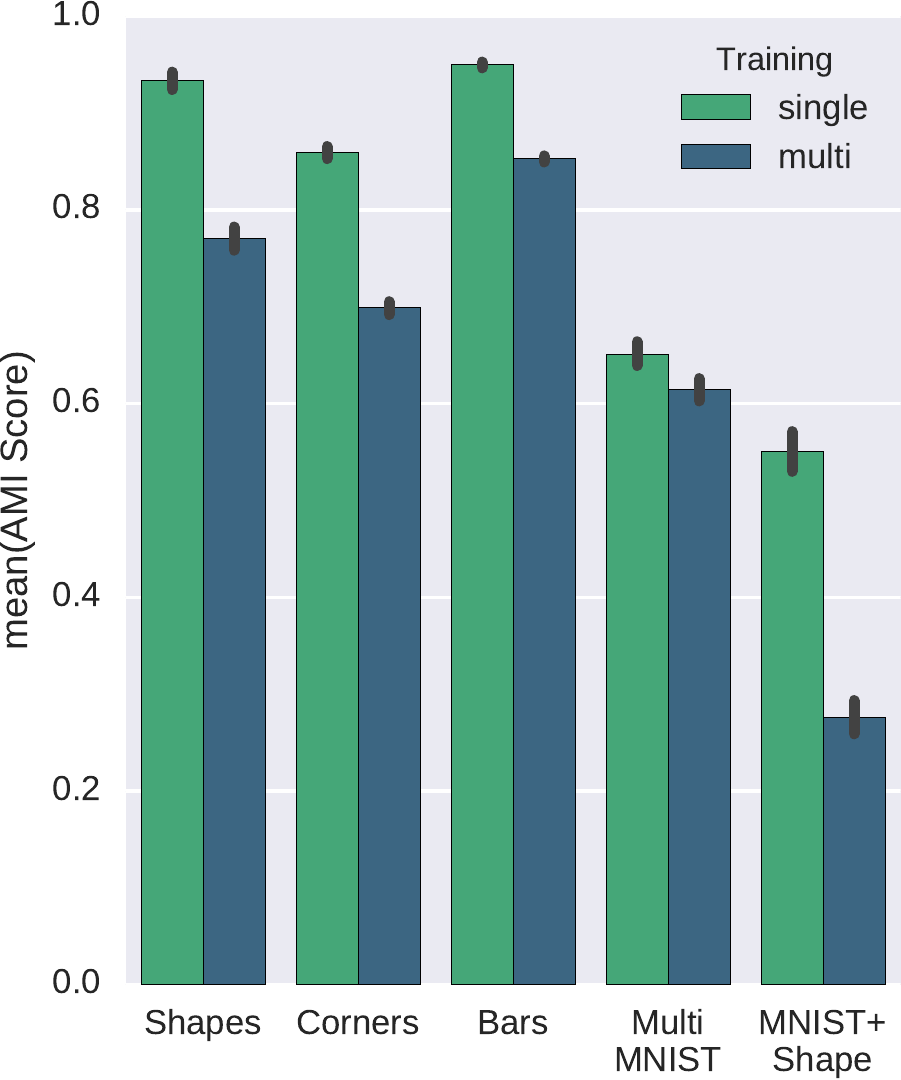}
    \caption{Single vs. Multi-object DAE training}
\label{fig:multi_vs_single}
    \end{subfigure}
    \caption{\textbf{Left:} Relationship between the DAE loss and the AMI score. All networks have 250 hidden units and were trained with random learning rates and initializations. A few networks that failed to train were removed from the plot for better visualization.
    \textbf{Right:} RC scores obtained when training DAEs on multi-object images vs. single object images.}
\end{figure}

RC utilizes autoencoders trained with the denoising objective for binding.
Therefore, it is instructive to examine the relationship between denoising performance and the final RC binding score.
For this purpose, we trained 100 DAEs with the same architecture on each dataset with random learning rates and initializations, and then performed RC using each of them.
\autoref{fig:score_vs_loss} shows the relationship between the denoising loss and binding score for each dataset.
It can be observed that lower loss correlates positively with higher score for all datasets, indicating that denoising is a suitable surrogate training objective.
We added a regression line to indicate that relation for each dataset, even though for \emph{MNIST+Shape} and \emph{Multi MNIST} it doesn't look even remotely linear.
Instead, the individual points are approximately arranged on a curve.
This suggests that there is a direct but complex interplay between the denoising performance and the score.

\subsection{Training on Multiple Objects}
So far the DAEs were trained on single-object images, then used to bind objects in multi-object images.
In general it is desirable to not \emph{require} single-object images for training, and be able to directly use any image without this restriction.
This would remove the last bit of supervision and make RC a truly unsupervised method. 

Why should this work at all? 
On the surface it seems that RC would depend on the DAE to prefer single objects in order to work correctly.
However, even if each cluster tries to reconstruct every object, there will be small asymmetries due to the difference in inputs they see.
Since no object carries any information about the shape and position of another object in our datasets, this will lead to differences in prediction quality of the objects. 
The resulting difference in reconstruction quality will then be amplified by RC and can still lead to a segregation of the objects.

To test this scenario, we performed a new random search to tune DAE hyperparameters for the case of multi-object training. 
Similar to the single-object case, we then used the best obtained DAEs to perform RC on test examples.
We found that with soft-assignments to the clusters, the differences were too small and would even out over several iterations, leading to uniform cluster assignments. 
By changing the E-step to hard (K-Means-like) assignments, we were able to amplify these changes enough to make the whole procedure work. 
\autoref{fig:multi_vs_single} shows that DAEs trained on multi-object images can indeed be used for binding via RC with hard assignments, although they lead to lower scores in comparison. Further discussion and examples for this case can be found in \autoref{sec:app_multi}.


\subsection{Generalization to a new domain}
\begin{figure}
\centering
\includegraphics[width=0.9\textwidth]{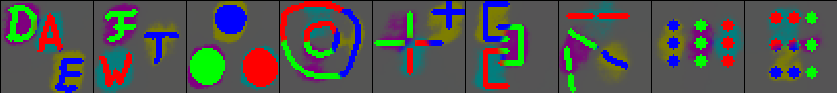}
\caption{Binding novel objects via RC. The DAE used was trained on the \emph{Multi MNIST} dataset.}
\label{fig:gestalt}
\end{figure}

A central intuition behind our approach to binding is that the low-level structures learned by the model will generalize to new and unseen configurations. 
Evaluation on unseen test sets demonstrated this to be true, but we can take it one step further.
We can test what happens when we confront our method with novel objects that the auto-encoders have not been trained on.

We ran RC on several images with non-digits using a DAE trained on the \emph{Multi-MNIST} dataset. 
\autoref{fig:gestalt} shows that RC ``correctly'' binds letters and circles together.
We also show images for which the resulting binding differs from our expectation.
It appears that the network has mainly learned to bind based on spatial proximity with a slight bias towards vertical proximity.
This can be expected since that it has only seen digits of roughly the same size so far, and because the used autoencoder is very limited. 
Nevertheless, it is very interesting that a fully-connected network which is permutation invariant learns the preference for spatial proximity entirely from data.
It is reasonable to speculate that it in the future it may be possible to recover other \emph{Gestalt Principles} such as continuity and similarity with a similar procedure.


\section{Relationship to other Methods}\label{sec:related}
The binding problem and its possible solutions are a long standing debate in the neuroscience literature (see e.g. \cite{milner1974, vonderMalsburg1981, Gray1999, Treisman1999, dilollo2012}).
A major thread of work on binding has been inspired by the temporal correlation theory \citep{vonderMalsburg1981}, based on utilizing synchronous oscillations to bind neuronal activities together. \citet{vondermalsburg1995} provides an overview of these ideas.
Recently, these ideas were implemented using complex valued activations in neural networks to jointly encode firing rate and phase \citep{Rao2008,Reichert2013}.
Such binding mechanisms are close to their biological inspiration, clustering only implicitly through synchronization.
In contrast, RC is based on a mathematical framework which explicitly incorporates binding.

Mechanisms for tackling the binding problem which do not require temporal synchronization have also been proposed (e.g. \citealp{oreilly2003}).
\citet{oreilly2002} argued that the intuitive explanation of the binding problem from \autoref{fig:interference} only applies if the distributed features themselves are local codes. 
They suggested that neural networks can avoid the binding problem using coarse-coded representations.
Various feature representation types including coarse-coding and their limitations were described by \citet{Hinton1984}.

In principle, Recurrent Neural Networks (RNNs; e.g.~\citealp{Robinson1987, Werbos1988}) can solve the binding problem by learning a mechanism to avoid it.
Psychologists \citep{dilollo2012} and machine learning researchers \citep{weng2006} alike have suggested feedback as a mechanism to do binding.
An RNN may utilize an implicit or explicit attention mechanism to selectively process different parts of the input \citep{schmidhuber1991,mnih2014,bahdanau2014}.
In this context, explicit binding via RC can be seen as a technique of paying attention to multiple objects at once, instead of focusing on them sequentially.

In some aspects, RC is similar to segmentation algorithms.
The main difference is that RC learns the segmentation from the data in a largely unsupervised manner.
In this sense, it is more similar to \emph{superpixel} methods (see e.g. \cite{achanta2012} for an overview).
However, these methods impose a handcrafted similarity measure over pixels or pixel regions, whereas RC learns a non-linear similarity measure from the data, parameterized by a DAE.


\section{Conclusion and Future Work}\label{sec:conclusion}
We introduced the Reconstruction Clustering framework to explicitly model data as a composition of objects, where the notion of object-ness is defined by mutual predictability. 
Compared to many previous solutions to the binding problem, this framework is mathematically rigorous, integrates well with current representation learning methods, and is effective for a variety of binary image datasets.
While a typical representation learning method (such as a denoising autoencoder) learns a \emph{static binding} of features, Reconstruction Clustering utilizes it to iteratively perform \emph{dynamic binding} for every input example by introducing interaction between the statically bound features extracted by the autoencoder.
In particular, this interaction enables dynamic binding of feature combinations never seen before by the autoencoder. 

This paper lays the groundwork for many concrete lines of future exploration.
The treatment of real-valued inputs is an important next step to extension RC towards natural data.
Also the use of more powerful autoencoders will be key. 
Integrating RC with the training of the DAE should help to deal with multiple objects in the training data. 
Since the method is general, we expect to apply it to other domains such as audio data (binding different speaker voices together) or medical data (binding various related symptoms of disease together).
A particularly interesting direction for future work is to show that Gestalt principles are a natural result of such a representation learning approach. 


\subsubsection*{Acknowledgments}
We thank Jan Koutn\'{i}k, Sjoerd van Steenkiste, Boyan Beronov and Julian Zilly for helpful discussions and comments. This project was funded by the EU project NASCENCE (FP7-ICT-317662).

\bibliography{Binding}
\bibliographystyle{iclr2016_conference}

\pagebreak
\appendix

\section{Reconstruction Clustering Derivation}
This section contains a more detailed derivation of Reconstruction Clustering (RC) for binary inputs. 
It follows the notation and derivation of an Expectation Maximization (EM) algorithm wherever possible.
Only for the M-step does RC deviate from EM.

Consider $N$ binary random variables (one for each pixel) that are distributed according to a mixture of $K$ Bernoulli distributions with means $\bm{\mu}_i = (\mu_{i1}, \mu_{i2}, \dots, \mu_{iK})$ and mixing coefficients $\bm{\pi}=(\pi_1, \pi_2, \dots, \pi_K)$ that sum to one $\sum_{k=1}^K \pi_k = 1$. Under this model the data likelihood given the parameters is given by: 

\begin{equation}\label{eq:px_mp}
    P(x_i|\bm{\mu}_i, \bm{\pi}) = \sum_{k=1}^K \pi_k \mu_{ik}^{x_i} (1-\mu_{ik})^{1-x_i}
\end{equation}

By defining $\mathbf{x} = (x_1, x_2, \dots, x_N)$ and $\bm{\mu} = (\bm{\mu}_1, \bm{\mu}_2, \dots, \bm{\mu}_N)$ and assuming independence of the $x_i$'s given $\bm{\mu}$ and $\bm{\pi}$ (but not identical distribution)\footnote{This assumption means that we assume the hidden representation of the DAE to capture the structure in the image well.} we get the (incomplete) log likelihood function for this model:

\begin{align}
    \log P(\mathbf{x}| \bm{\mu}, \bm{\pi}) &= \sum_{i=1}^N \log P(x_i|\bm{\mu}_i, \bm{\pi})
\end{align}

Let us now introduce an explicit binary latent variable $\mathbf{z_i} = (z_{i1}, z_{i2}, \dots, z_{iK}) \in \{0, 1\}^K$ with $\sum_{k=1}^K z_{ik} = 1$ associated with each $x_i$. 
Let the prior distribution be:

\begin{align}
    P(\mathbf{Z}|\bm{\pi}) &= \prod_{i=1}^N P(\mathbf{z}_i|\bm{\pi}) = \prod_{i=1}^N \prod_{k=1}^K \pi_k^{z_{ik}},
\end{align}

where we set $\mathbf{Z}= (\mathbf{z}_1, \mathbf{z}_2, \dots, \mathbf{z}_N)$ and assume $\mathbf{z}_i$'s to be independent given $\bm{\pi}$.
With that we define the conditional distribution of $x_i$ given the latent variables as:

\begin{align}
    P(x_i |\mathbf{z}_i, \bm{\mu}_i) &= \prod_{k=1}^K P(x_i|\mu_{ik})^{z_{ik}}\\
    \label{eq:px_zmp}
    &= \prod_{k=1}^K (\mu_{ik}^{x_i} (1-\mu_{ik})^{1-x_i})^{z_{ik}}
\end{align}

If we marginalize \autoref{eq:px_zmp} over all choices of $\mathbf{z}_i$ we recover \autoref{eq:px_mp}:

\begin{align}
    \sum_{\mathbf{z}} P(x_i |\mathbf{z}, \bm{\mu}_i, \bm{\pi}) P(\mathbf{z}|\bm{\pi}) &=  \sum_{\mathbf{z}} \prod_{k=1}^K (\mu_{ik}^{x_i} (1-\mu_{ik})^{1-x_i})^{z_{k}} \pi_k^{z_{k}}\\
    &=\sum_{k=1}^K \pi_k \mu_{ik}^{x_i} (1-\mu_{ik})^{1-x_i}\\
    &= P(x_i|\bm{\mu}_i, \bm{\pi})
\end{align}

The second line is obtained by realizing that the $\sum_\mathbf{z}$ sums over exactly $K$ terms, each corresponding to a $\mathbf{z}$ with one $z_k=1$ and all other entries equal to zero. 
So we can replace this sum by $\sum_{k=1}^K$.
The product over the entries of $\mathbf{z}$ then vanishes except for the term corrsponding to the $k$-th entry.

Using the same conditional independence assumption from before we can thus write the data distribution given all the latent variables as follows: 

\begin{align}
    P(\mathbf{x}| \mathbf{Z}, \bm{\mu}, \bm{\pi}) &= \prod_{i=1}^N P(x_i|\mathbf{z}_i, \bm{\mu}_i, \bm{\pi})
\end{align}

And by using Bayes rule and assuming that $\mathbf{Z}$ is independent of $\bm{\mu}$:

\begin{align}
    P(\mathbf{x}, \mathbf{Z}| \bm{\mu}, \bm{\pi}) &= P(\mathbf{x}| \mathbf{Z}, \bm{\mu}, \bm{\pi}) P(\mathbf{Z} | \bm{\pi})\\
    &= \prod_{i=1}^N \prod_{k=1}^K (\mu_{ik}^{x_i} (1 - \mu_{ik})^{1-x_i} \pi_k)^{z_{ik} } 
\end{align}

If we set $\bm{\theta} = \{\bm{\mu}, \bm{\pi}\}$\footnote{Here we deviate slightly from the notation in the paper.} the complete-data log likelihood becomes:

\begin{align}
    \mathcal{L}(\bm{\theta}| \mathbf{x, Z}) &= \log P(\mathbf{x}, \mathbf{Z}| \bm{\mu}, \bm{\pi})\\
     &= \sum_{i=1}^N \sum_{k=1}^K z_{ik} \left[ x_i \log \mu_{ik} + (1-x_i) \log (1 - \mu_{ik}) + \log \pi_k\right]
\end{align}

To maximize $\mathcal{L}$ with respect to $\bm{\theta}$ and $\mathbf{Z}$ we follow the same idea as the EM algorithm:
Based on the observation that if we knew the values of either of these two, optimizing the other would be feasible. 
So we divide the optimization problem into two steps where we pretend to know either $\bm{\theta}$ (E-step) or $\mathbf{Z}$ (M-step).

In the E-Step we assume to know $\bm{\theta}$ and  calculate the posterior probability of $z_{ik} = 1$ for each datapoint calling it $\gamma_{ik}$: (We assume the $z_{ik}$ to be independent of $x_j$ for $i\ne j$)
\begin{align}
    \gamma_{ik} = P(z_{ik}=1|x_i, \bm{\mu_i}, \bm{\pi}) &= \ddfrac
       {P(x_i|z_{ik}=1, \bm{\mu_i}) P(z_{ik}=1|\bm{\pi})}
       {P(x_i|\bm{\mu_i}, \bm{\pi})}\\[5pt]
    &= \ddfrac
       {\mu_{ik}^{x_i} (1-\mu_{ik})^{1-x_i} \pi_k }
       {\sum_{j=1}^K \mu_{ij}^{x_i} (1-\mu_{ij})^{1-x_i} \pi_j }
\end{align}

Next we calculate the $\mathcal{Q}$ value used in EM which is defined as the expectation of the complete data log-likelihood $\mathcal{L}$ with respect to the posterior of $\mathbf{Z}$ given the data and the old parameters $\bm{\theta}^{old}$:

\begin{align}
    \mathcal{Q}(\bm{\theta}, \bm{\theta}^{old}) &= E_{\mathbf{Z}}[\log P(\mathbf{x}, \mathbf{Z}| \bm{\theta}) | \mathbf{x}, \bm{\theta}^{old}]\\
     &= \sum_{\mathbf{Z}} P(\mathbf{Z}|\mathbf{x}, \bm{\theta}^{old})\log P(\mathbf{x}, \mathbf{Z}| \bm{\theta})\\
     &= \sum_{i=1}^N \sum_{k=1}^K \gamma_{ik} \left[ x_i \log \mu_{ik} + (1-x_i) \log (1 - \mu_{ik}) + \log \pi_k\right]
\end{align}

In the M-step of EM we aim to maximize $\mathcal{Q}(\bm{\theta}, \bm{\theta}^{old})$ over all choices of $\bm{\theta}$:

\begin{equation}
  \bm{\theta}^{new} = \argmax_{\bm{\theta}} \mathcal{Q}(\bm{\theta}, \bm{\theta}^{old})
\end{equation}

Using a Lagrange multiplier to enforce $\sum_{k=1}^K \pi_k = 1$ we find:

\begin{equation}
    \pi_k^{new} = \frac{\sum_{i=1}^N \gamma_{ik}}{N}
\end{equation}

But when maximizing wrt. $\bm{\mu}$ we see that the maximum is trivially obtained by setting $\mu_{ik} = x_i$ for all $k$. This is due to the fact that the problem is actually ill-posed in the sense that we have $K$ parameters to fit for each datapoint. So there are infinitely many solutions which achieve the optimal log likelihood of the data of 0. 

At this point we introduce an autoencoder with encoder $f$ and decoder $g$ to restrict the capacity of our model by forcing $\mu$ to be:
\begin{equation}
    \bm{\mu}_{\cdot k} = g(f(\bm{\gamma}_k \odot \mathbf{x}))
    \label{eq:r_step}
\end{equation}

We use this reconstruction step (\autoref{eq:r_step}) instead of an actual maximization step, thus deviating from the EM formulation.

\section{Training Details}
All experiments have been performed with the \href{https://github.com/IDSIA/brainstorm}{brainstorm} library and were organized and logged using \href{https://github.com/IDSIA/Sacred}{sacred}.
The code for this paper can be found on \href{https://github.com/Qwlouse/Binding}{GitHub}.
\subsection{Training Denoising Autoncoders}
\begin{itemize}
    \item simple feed forward fully connected NNs
    \item with sigmoid output layer 
    \item loss is Binomial Cross Entropy Error
    \item trained with SGD
    \item minibatch size 100
    \item salt\& pepper noise
    \item early stopped when validation BinomialCEE doesn't decrease for more than 10 epochs
\end{itemize}

\subsection{Random Search}
There are several hyperparameters to be chosen for the denoising autoencoders. 
To find good values we performed a random search with 100 runs for each dataset.
For each run we randomly sampled from the following parameters:
\begin{itemize}
    \item learning rate log-uniform from $[10^{-3}, 1]$
    \item Amount of Salt\& Pepper Noise from $[0.0, 0.1, \dots, 0.9]$
    \item hidden layer size from $[100, 250, 500, 1000]$
    \item hidden layer activation function from [rel, sigmoid, tanh]
\end{itemize}
The best network configurations found by that search can be found in \autoref{tab:best_net}.

\begin{figure}[h]
\centering
\includegraphics[width=0.75\textwidth]{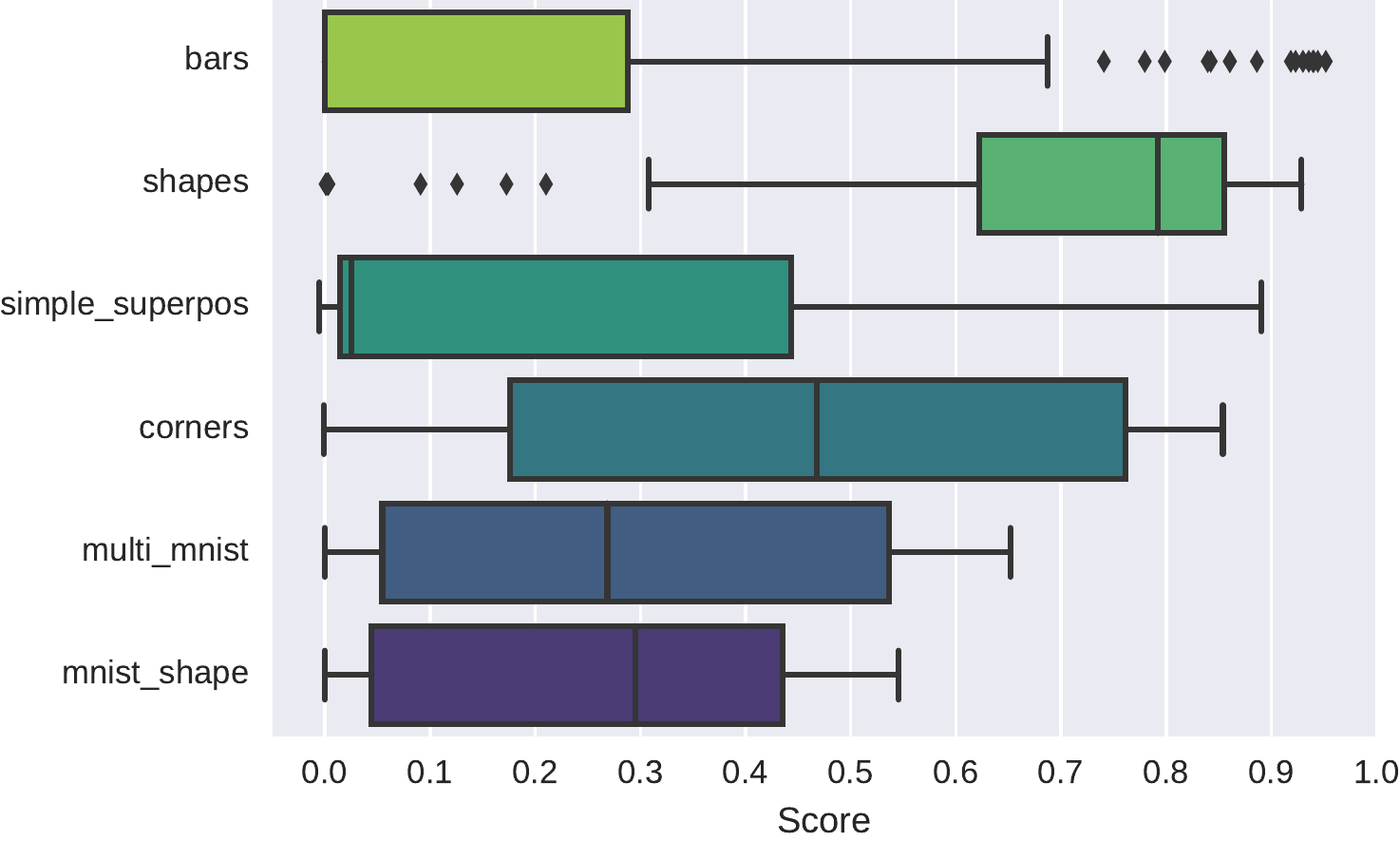}
\caption{Summary of the scores achieved during the random search}
\label{fig:random_search}
\end{figure}

\begin{table}[h]
\centering
\begin{tabular}{lrcccr}
\toprule
Dataset &  learning rate & \# hidden units & activation & salt\&pepper &     score \\
\midrule
    bars &       0.768015 &        100 &  ReL    &      0.0 &  0.951809 \\
    corners &       0.001920 &        100 &  ReL    &      0.0 &  0.853866 \\
    multi\_mnist &       0.011362 &       1000 & ReL  &         0.6 &  0.651657 \\
    mnist\_shape &       0.031685 &        250 &  sigmoid  &        0.6 &  0.545559 \\
    shapes &       0.083147 &        500 &   tanh   &      0.4 &  0.928792 \\
    simple\_superpos &       0.366627 &        100 &   ReL  &       0.1 &  0.890472 \\
\bottomrule
\end{tabular}
\caption{Configuration of the best network for each dataset as found by the random search.}
\label{tab:best_net}
\end{table}

\subsection{Random Search for Training with Multiple Objects}
For training with multiple objects we do an equivalent random search for hyperparameters. 
The only difference is the training data and that for determining the final score we use K-means-like (hard) cluster assignments in RC. 
Note also that we didn't include the Simple Superposition dataset, since we only have 120 images with multiple objects available, and no separate test set.

\begin{figure}[h]
\centering
\includegraphics[width=0.75\textwidth]{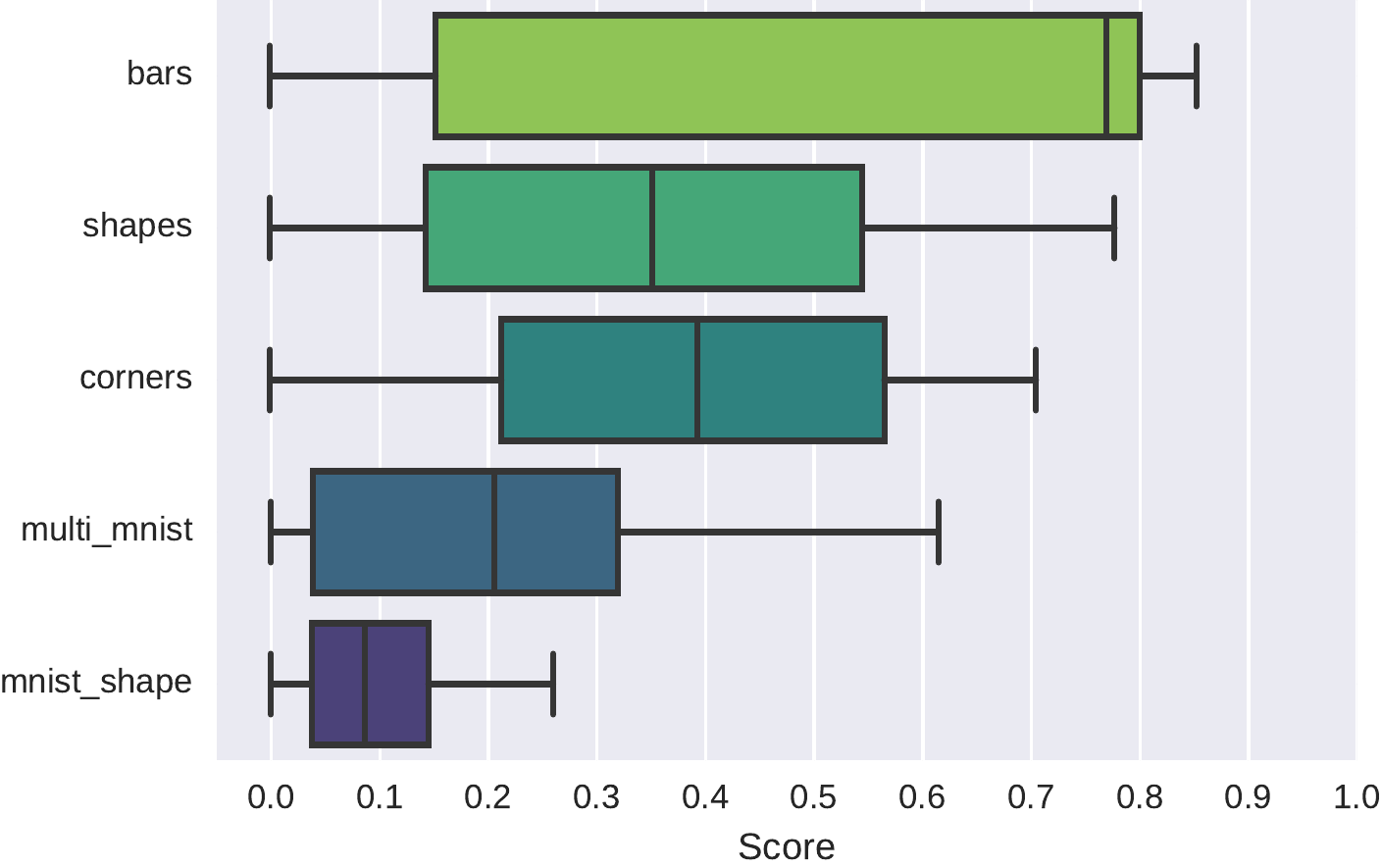}
\caption{Summary of the scores achieved during the random search for training with multiple objects}
\label{fig:random_search_multi}
\end{figure}

\begin{table}[h]
\centering
\begin{tabular}{lrcccr}
\toprule
Dataset & learning rate & \# hidden units & activation & salt\&pepper &     score \\
\midrule
    bars &       0.012192 &        100 & sigmoid &          0.8 &  0.851777 \\
    corners &       0.026035 &        100 & ReL &        0.7 &  0.704285 \\
    mnist\_shape &       0.033200 &       1000 & ReL &          0.6 &  0.259646 \\
    multi\_mnist &       0.001786 &        250 &  sigmoid &         0.9 &  0.614277 \\
    shapes &       0.049402 &        100 & sigmoid    &       0.9 &  0.776656 \\
\bottomrule
\end{tabular}
\caption{Configuration of the best network trained on \emph{multiple objects} for each dataset as found by the random search.}
\label{tab:best_multi_net}
\end{table}

\clearpage

\section{Multi Object Training}
\label{sec:app_multi}
When training the DAEs on images with multiple objects, it is less obvious why running RC should lead to a segregation of the objects. 
It seems that the autoencoder should always try to reconstruct the whole image including all the objects.
And if we run normal (soft) RC we in fact see that after a few iterations each pixel is equally represented by each cluster.

\begin{figure}[t]
\centering
\includegraphics[width=\textwidth]{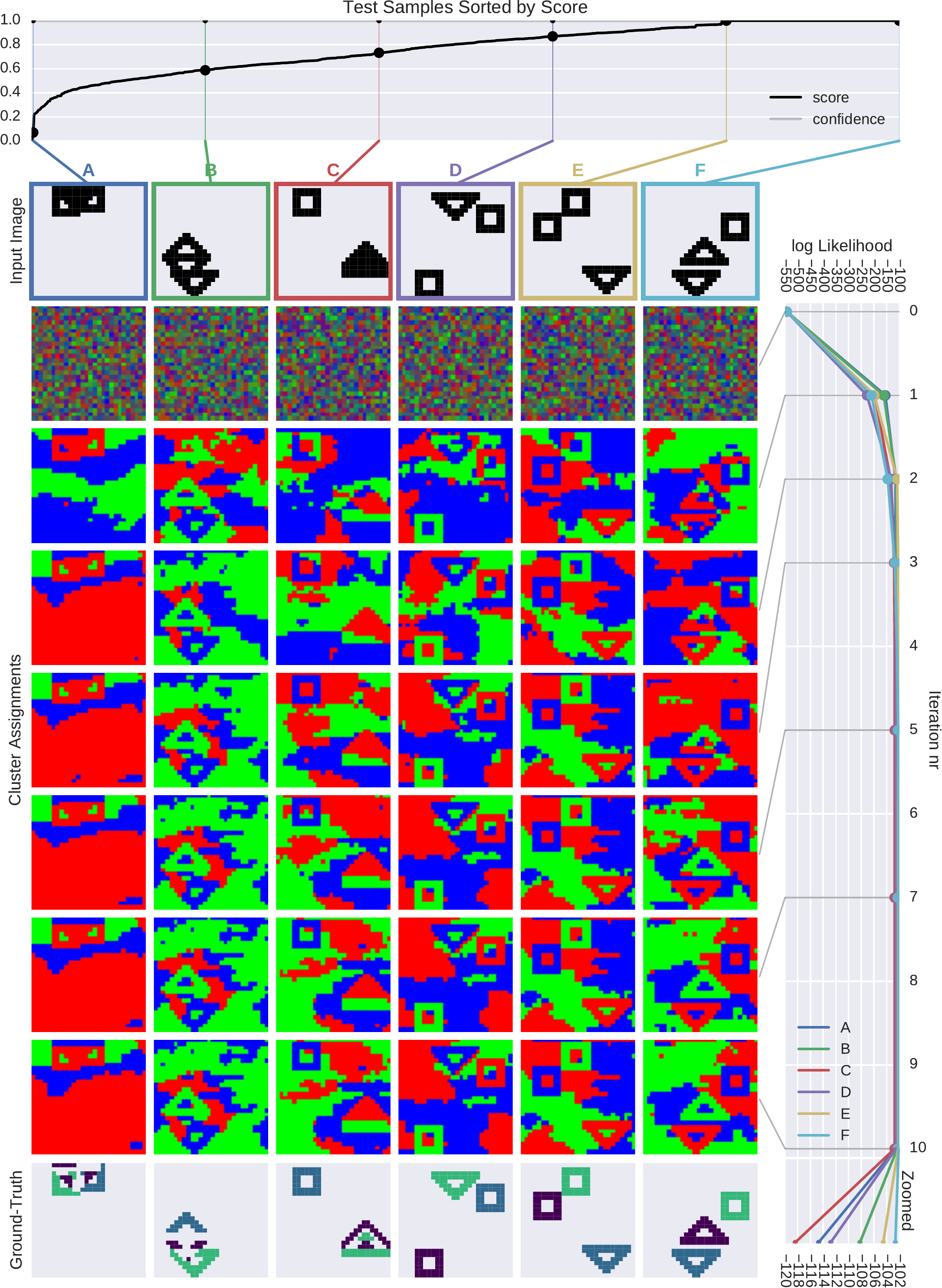}
\caption{Example iterations of RC when using hard assignments and a DAE that has been trained only on images with multiple objects.}
\label{fig:shapes_iter_multi_train}
\end{figure}

By switching to hard cluster assignments we eliminate this stable state, and force the clusters to compete more for the pixels. 
Together with the fact that in our datasets objects don't carry any information about other objects this leads to a stronger amplification of the initial differences in reconstruction quality.
In \autoref{fig:shapes_iter_multi_train} this process can be seen on the shapes dataset.
Note that the hard RC converges even faster, but generally leads to worse performance.

\section{Additional Figures}\label{sec:add-figures}

\begin{figure}[h]
\centering
\includegraphics[width=\textwidth]{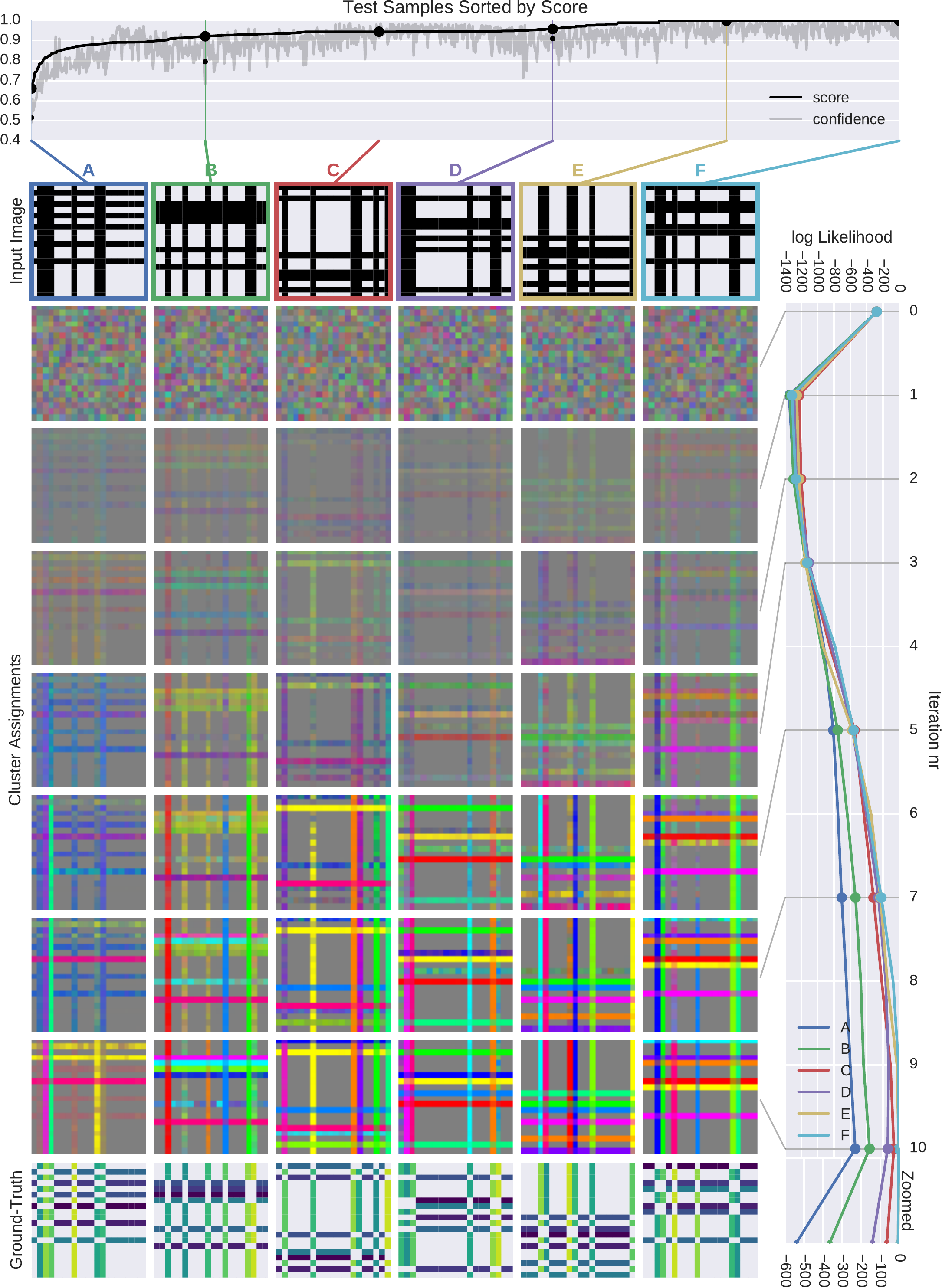}
\caption{}
\label{fig:bars_iter}
\end{figure}

\begin{figure}[p]
\centering
\includegraphics[width=\textwidth]{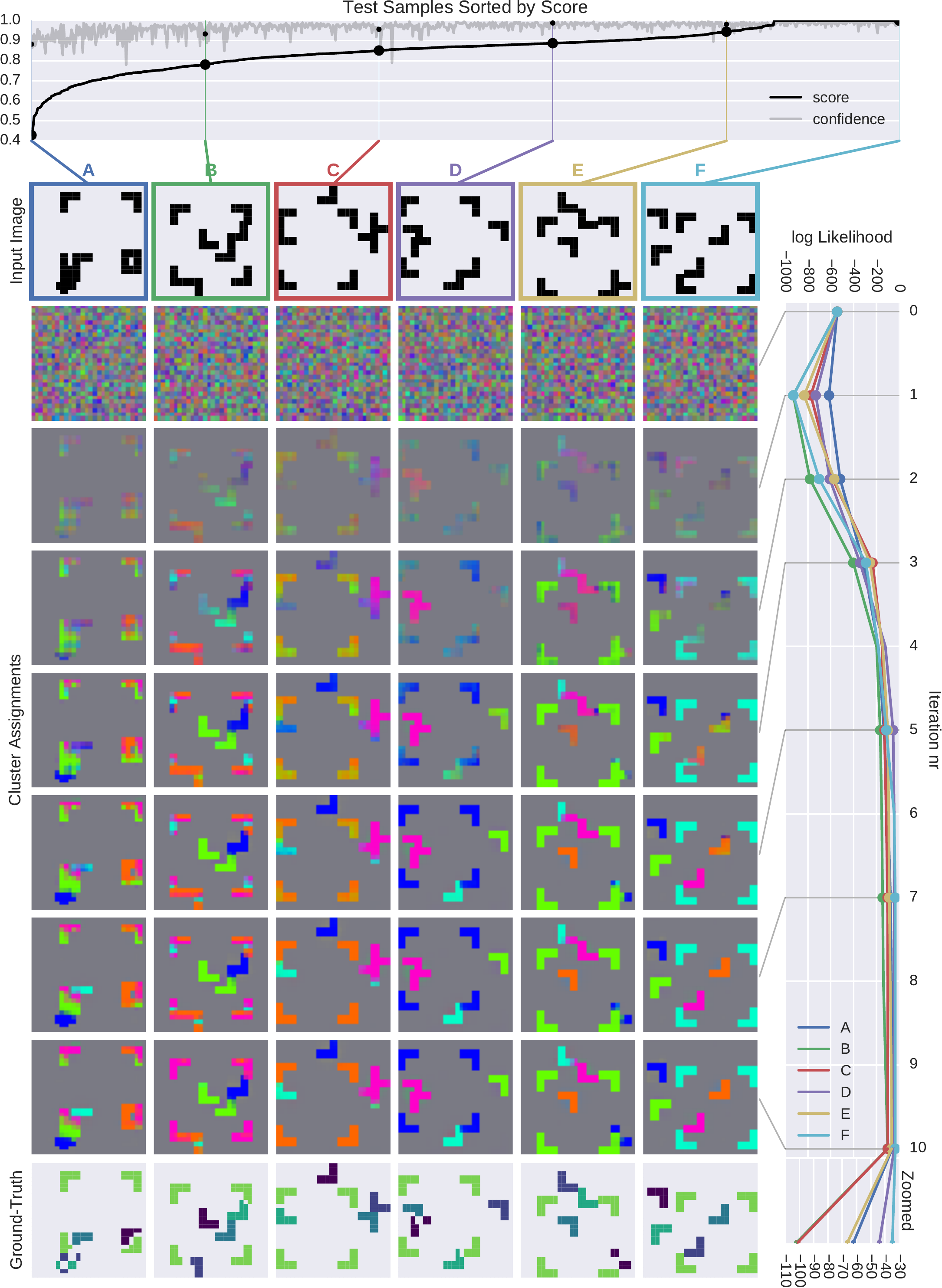}
\caption{}
\label{fig:corners_iter}
\end{figure}

\begin{figure}[t]
\centering
\includegraphics[width=\textwidth]{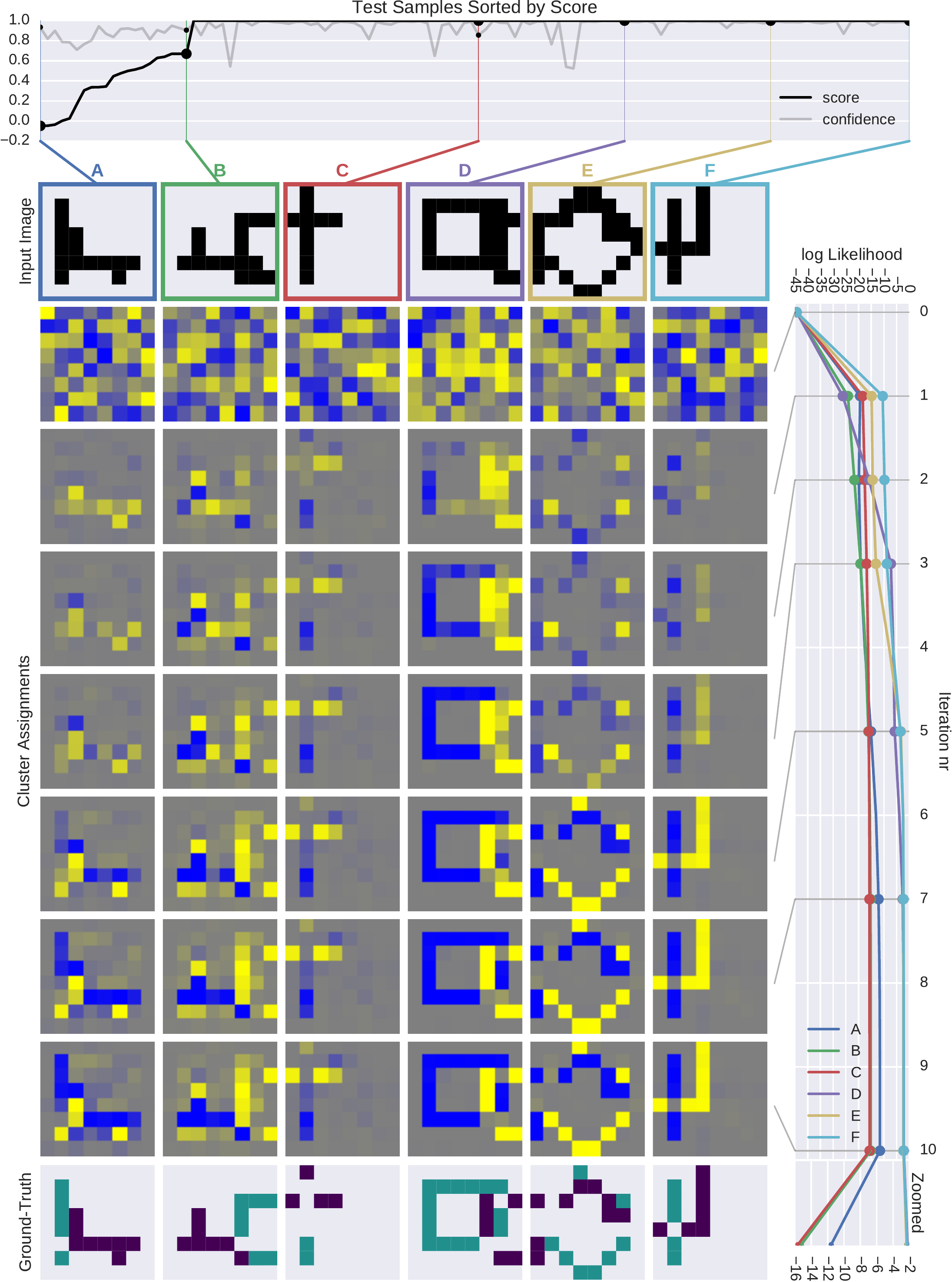}
\caption{}
\label{fig:superpos_iter}
\end{figure}

\begin{figure}[t]
\centering
\includegraphics[width=\textwidth]{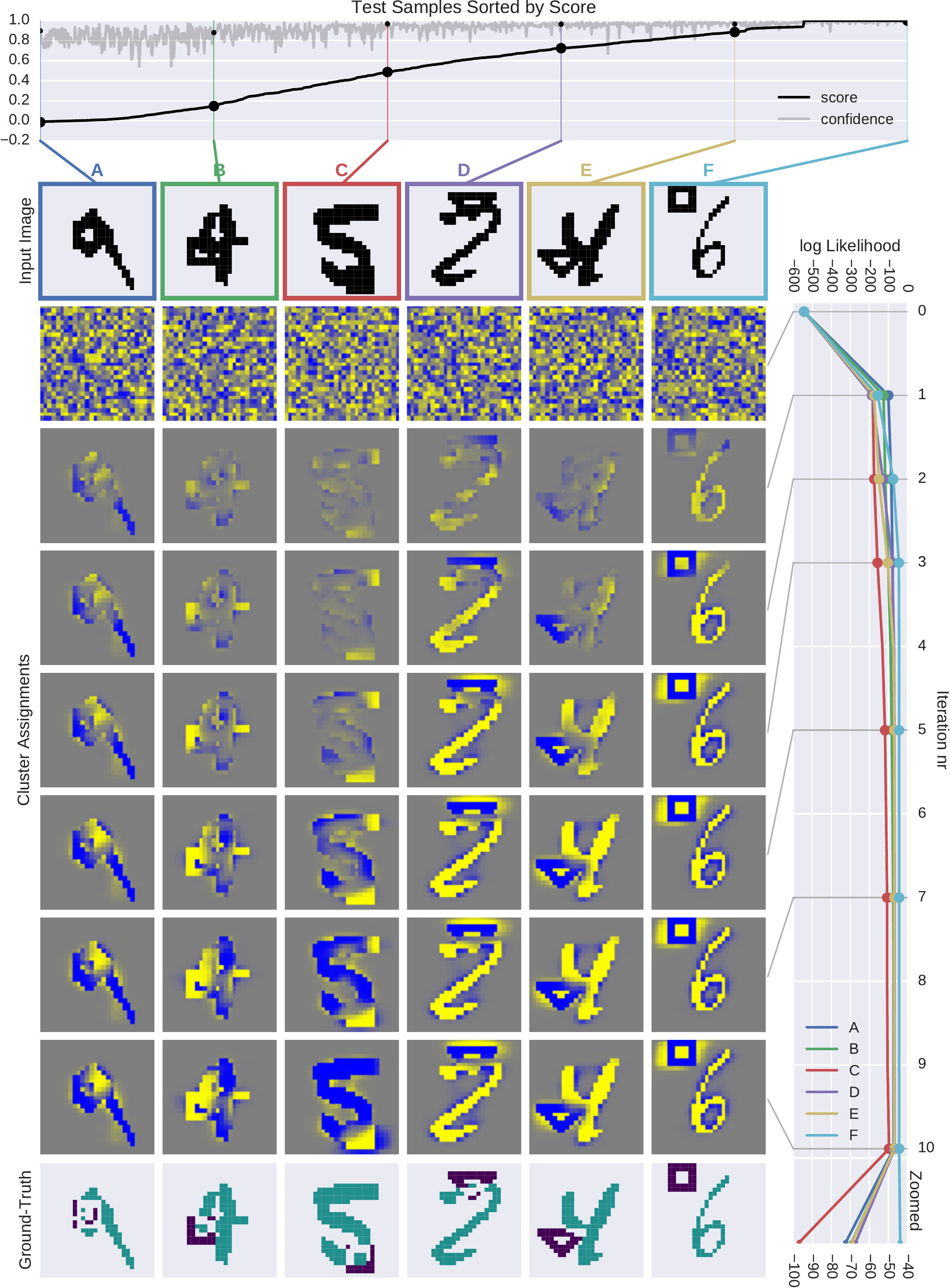}
\caption{}
\label{fig:mnist_shapes_iter}
\end{figure}

\begin{figure}[t]
\centering
\includegraphics[width=\textwidth]{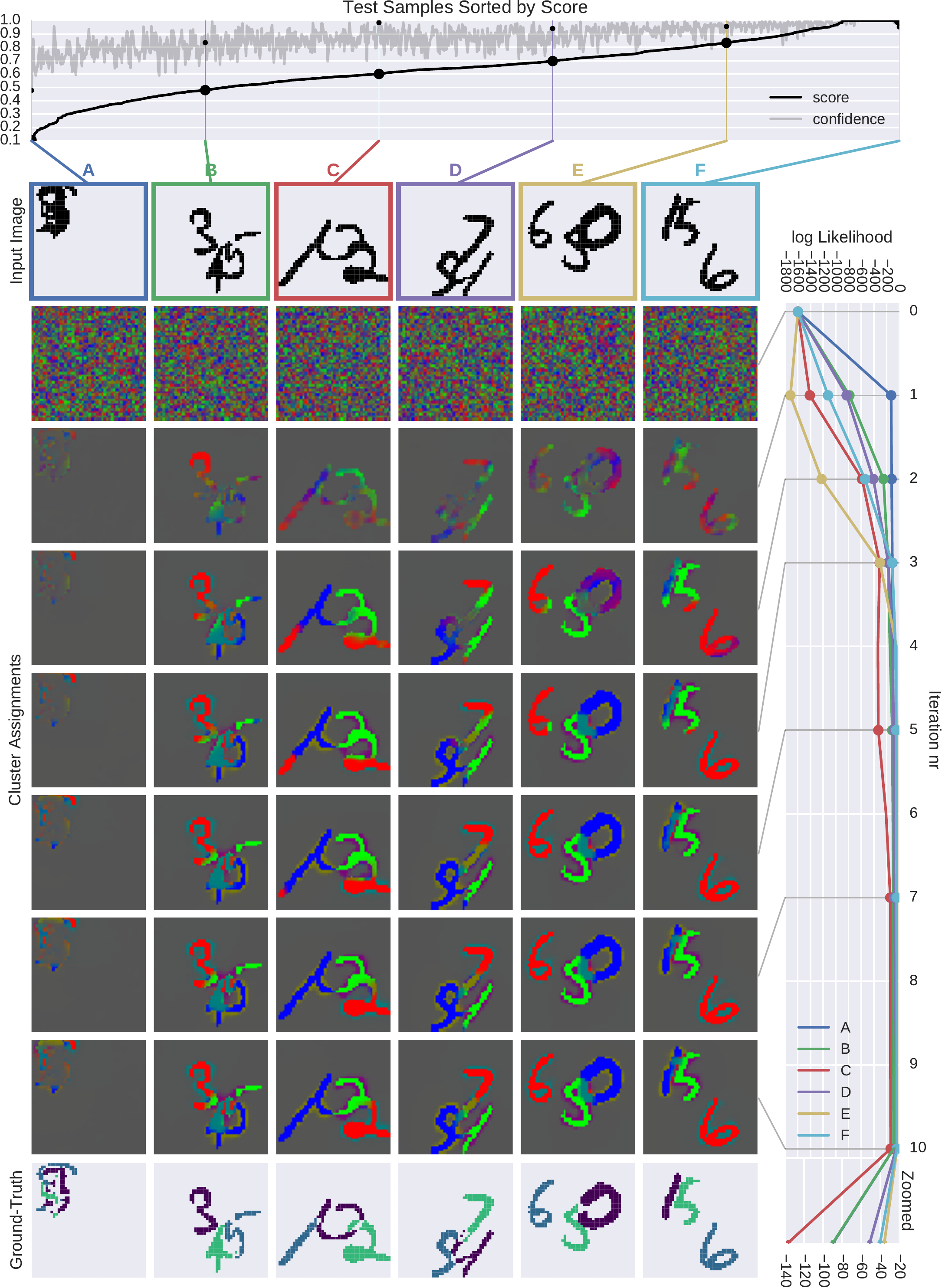}
\caption{}
\label{fig:multi_mnist_iter}
\end{figure}

\end{document}